\documentclass{article}

\usepackage[preprint]{neurips_2025}

\usepackage[utf8]{inputenc} 
\usepackage[T1]{fontenc}    
\usepackage{hyperref}       
\usepackage{url}            
\usepackage{booktabs}       
\usepackage{amsfonts}       
\usepackage{nicefrac}       
\usepackage{microtype}      
\usepackage{xcolor}         
\usepackage{graphicx}
\usepackage{subcaption} 

\usepackage{amsmath}  
\usepackage{xspace}
\usepackage{hyperref}
\usepackage{xcolor,cleveref}
\usepackage{graphicx}  
\usepackage{pifont}  
\usepackage{marginnote}
\usepackage{algorithm}
\usepackage{algpseudocode}
\usepackage{multirow}
\usepackage{tcolorbox}
\usepackage{enumitem}
\usepackage{amsthm}

\usepackage{tabularray}  

\newif\ifsubmit
\submitfalse

\newcommand{\remove}[1]{}

\newcommand{\reviewer}[3]{
  \expandafter\newcommand\csname #1\endcsname[1]{
    \ifsubmit
      \ignorespaces
    \else
      \textcolor{#3}{[#2: ##1]}
    \fi
  }
}

\reviewer{nini}{Ni}{purple}
\reviewer{brafalsk}{brafalsk}{purple}
\reviewer{xiaoyuan}{Xiaoyuan}{orange}
\reviewer{karen}{Karen}{blue}
\reviewer{john}{John}{green}
\reviewer{alex}{Alex}{blue}

\tcbset{
    sharp corners,
    colback = white,
    before skip = 0.2cm,    
    after skip = 0.5cm      
}                           

\definecolor{main}{HTML}{5989cf}    
\definecolor{sub}{HTML}{cde4ff}     
\newtcolorbox{prompt}{
    colback = sub,
    boxrule = 0pt, 
    colframe = white, 
}

\title{
Can LLMs Ask Good Questions?
}

\author{
  \textbf{
    Yueheng Zhang\textsuperscript{1}\thanks{These authors contributed equally to this work.}, 
    Xiaoyuan Liu\textsuperscript{1}\footnotemark[1], 
    Yiyou Sun\textsuperscript{1}, 
    Atheer Alharbi\textsuperscript{2}, 
  } \\
  \textbf{
    Hend Alzahrani\textsuperscript{2}, 
    Tianneng Shi \textsuperscript{1},
    Basel Alomair\textsuperscript{2,3}, 
    Dawn Song\textsuperscript{1}
  } \\
  \textsuperscript{1} University of California Berkeley, 
  \textsuperscript{2} KACST,
  \textsuperscript{3} University of Washington \\
  \texttt{\{azicon, xiaoyuanliu, sunyiyou, stneng, dawnsong\}@berkeley.edu}, \\
  \texttt{\{asalharbi, hmmalzahrani\}@kacst.edu.sa}, 
  \texttt{alomair@uw.edu}
}

\begin{document}

\maketitle
\begin{abstract}

We evaluate questions generated by large language models (LLMs) from context, comparing them to human-authored questions across six dimensions: question type, question length, context coverage, answerability, uncommonness, and required answer length.  
Our study spans two open-source and two proprietary state-of-the-art models. Results reveal that LLM-generated questions tend to demand longer descriptive answers and exhibit more evenly distributed context focus, in contrast to the positional bias often seen in QA tasks. These findings provide insights into the distinctive characteristics of LLM-generated questions and inform future work on question quality and downstream applications.

\end{abstract}


\section{Introduction}



    







People have long designed questions with known answers for various purposes, such as education, dialog systems, and model evaluation~\citep{guo2024survey}. For automation purposes, Question Generation (QG) is a task focused on creating relevant questions based on given facts. Numerous automatic QG methods have been proposed in the literature~\citep{duan-etal-2017-question,pan2019recentadvancesneuralquestion}. Recent advances in large language models (LLMs) have greatly improved the performance of NLP tasks, including QG~\citep{es2023ragas}. 

Despite the widespread use of LLMs in QG, few in-depth studies have explored the characteristics of LLM-generated questions. Do LLMs, without additional prompt constraints, prefer longer or shorter questions? What types of questions do they tend to ask? How do LLM-generated questions differ from those created by humans? While existing research~\citep{es2023ragas} evaluates human alignment using empirical methods, there is a lack of comparison studies between LLM-generated questions and human-generated ones. A deeper understanding of LLM behavior in QG would provide valuable insights for prompt optimizations.

Many evaluation metrics and quality criteria have been proposed for the QG task~\citep{Gorgun2024,guo2024survey}, which can be categorized into two types. Some rely on statistical measures but fail to capture rich evaluation semantics, while others require extensive human labeling. 
With the increasing capabilities of LLMs, recent work~\citep{fu2023gptscore} has demonstrated the feasibility of automatic evaluation with richer semantics. This suggests new opportunities for more comprehensive evaluation with reduced human effort, which we explore in this study.

\begin{figure*}[hbt]
    \centering
    \includegraphics[width=\textwidth]{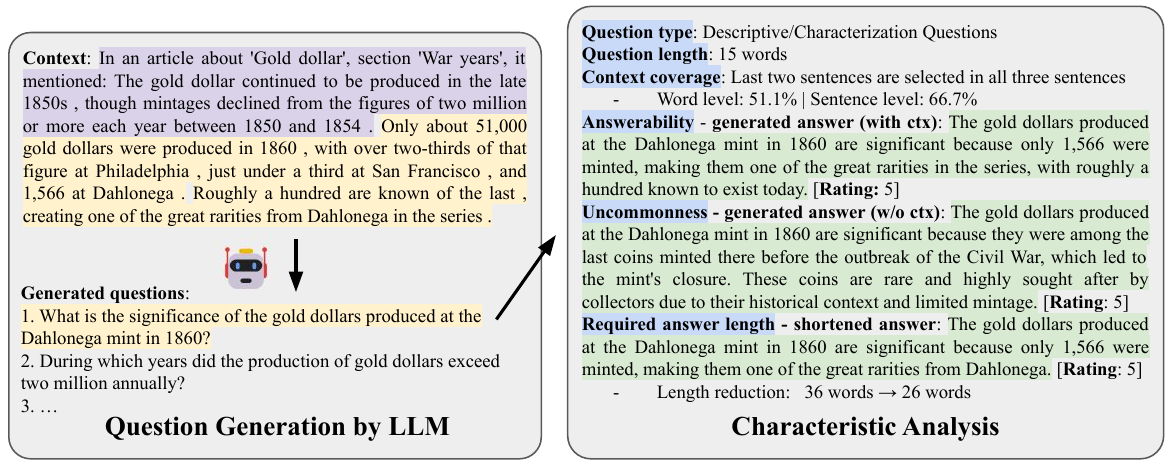}
    \caption{\label{fig:intro}Investigation on the characteristics of LLM-generated questions.}
\end{figure*}

In this paper, we evaluate the characteristics of LLM-based QG. Specifically, we compare LLM-generated questions with human-generated ones from a Wikipedia corpus, assessing six evaluation dimensions, as shown in Figure \ref{fig:intro}. Our findings reveal that LLMs tend to generate questions that require descriptive, longer answers. Additionally, unlike the positional bias often observed in QA~\citep{saito2024answerinvestigatingpositionalbias}, LLMs exhibit a more balanced focus across the entire context in QG.

Despite extensive research on LLM QA capabilities, this is, to our knowledge, the first study to uncover LLM preferences in QG. It also extends existing statistical question quality criteria by introducing an automatic evaluation workflow. Our findings provide valuable insights for evaluating downstream applications, such as Retrieval-Augmented Generation (RAG) systems and hallucination detection.

\section{Related work}
\label{sec:related}

In this section, we review prior works relevant to our study. 
We discuss existing question generation (QG) methods, their evaluation metrics, and recent advancements involving LLMs. 
Finally, we examine commonly used QA datasets and how their questions are constructed for comparison.





\paragraph{Question generation.}
QG~\citep{duan-etal-2017-question,guo2024survey,pan2019recentadvancesneuralquestion} aims to automatically produce a vast array of human-like high-quality questions. 
Different QG approaches utilize diverse sources for question content, including 
knowledge bases~\citep{knowbasedQGSPARQL,reddy-etal-2017-generating,knowledgeBasedGraph}, 
unstructured text~\citep{HUANG_HE_2016,TextQGTopictoQ, serban2016generatingfactoidquestionsrecurrent} 
and images~\citep{Li_2018_CVPR,mostafazadeh2016generating,knowledgeBasedVisual}.
Our work, similar to \citet{rajpurkar2016squad100000questionsmachine,JoshiTriviaQA2017,yang2018hotpotqa,kwiatkowski2019natural}, utilize Wikipedia-based knowledge.
\citet{guo2024survey} introduces a taxonomy of QG techniques. This taxonomy classifies QG methods based on the type of model used, where our solutions fall under the category of LLM-based QG.


\paragraph{QG quality evaluation metrics.}
There are many metrics used for evaluating QG quality. 
\citet{Gorgun2024} establishes a taxonomy for evaluation methods, providing a structured approach to assess QG systems. It categorizes evaluation methods into two main types: judgment-based and statistical-based that can be automated. 
Following this taxonomy, the QG evaluation in our work falls under the category of statistical post-hoc method, similar to \citet{lin-2004-rouge,BLUE,zhong-etal-2022-towards-uniEval,fu2024qgevalbenchmarkquestiongeneration}.
Besides, for evaluating the answerability of the generated questions, exact match (EM) is a commonly used approach for existing datasets as the answers are often as simple as an entity name. There have also been previous practices \cite{wang2024evaluating} in using LLM to evaluate QA answer quality. In this work, we extend this approach to generate regression-style scores to provide finer-grain judgements on the answer quality. 



\paragraph{LLM-based data generation.}

Recent studies~\citep{wang2023selfinstructaligninglanguagemodels, das2024surface, hartvigsen-etal-2022-toxigen, NEURIPS2023_9cb2a749, kamalloo2023hagridhumanllmcollaborativedataset, ji2023beavertails, yin2023dynosaurdynamicgrowthparadigm, eldan2023tinystoriessmalllanguagemodels, benallal2024cosmopedia} show the growing use of LLMs in dataset creation for various applications.
Self-instruct~\citep{wang2023selfinstructaligninglanguagemodels} emerged as a mainstream approach for creating instruction-based datasets. By leveraging iterative bootstrapping, Self-instruct allows models to generate instruction-following data without extensive manual annotation, outperforming manually curated datasets with fewer tasks.
Other works similarly utilize LLMs to generate diverse forms of data but incorporate additional mechanisms to ensure quality. ToolQA~\citep{NEURIPS2023_9cb2a749} and HAGRID~\citep{kamalloo2023hagridhumanllmcollaborativedataset}, for instance, combine automated generation with human-guided refinement steps to ensure accuracy, making use of human intervention in specific phases of dataset curation. Toxigen~\citep{hartvigsen-etal-2022-toxigen} employs a similar strategy through classifier-in-the-loop decoding to filter LLM-generated toxic and benign text, producing a balanced dataset for toxicity detection.
Synthetic datasets like Cosmopedia~\citep{benallal2024cosmopedia}, BeaverTails~\citep{ji2023beavertails} and Tinystories~\citep{eldan2023tinystoriessmalllanguagemodels} emphasize large-scale data generation with minimal human involvement, highlighting the scalability of LLM-generated datasets.
In our work, we generate questions in free form and incorporate answerability evaluations for a further validation.

\paragraph{Open-domain QA dataset.}
There have been many open-domain QA datasets~\citep{yang2018hotpotqa} created to evaluate NLP models. 
TriviaQA~\citep{JoshiTriviaQA2017} introduce a dataset with enthusiast-collected trivia questions with context.
HotpotQA~\citep{yang2018hotpotqa} generates QA pairs from Wikipedia text with crowd-sourcing efforts. 
Natural Questions~\citep{kwiatkowski2019natural} presents questions annotated from google search. 
All these datasets requires human efforts during their construction. In this work, we focus on using LLM to automate the QG procedure.

\section{Methodology}
\label{sec:methodology}


This section outlines our method for generating questions from WikiText contexts.
We also present six quantitative metrics used to evaluate and analyze the preferences of LLMs.

\subsection{Question generation from context}
Given a paragraph string as the context $C$ and a QG instruction prompt $P$ that asks to generate $N$ questions, a LLM $M$ output $N$ questions $Q$ that can be answered with the facts from the context. 
In expectation of generating self-contained, independently answerable questions, we mainly use the following system prompt in our setup while also evaluate other paraphrases mentioned in Appendix~\ref{sec:diffprompt}:

\begin{prompt}
\textbf{Question generation.}
Generate [N] self-contained questions based on the following content in an ordered list.
\end{prompt}


To construct the context $C$ for concrete evaluations, we adapt the WikiText dataset for its broad coverage in topics and wide use in existing works~\citep{yang2018hotpotqa,JoshiTriviaQA2017}.
For preprocessing, we split it into 860k paragraphs while preserving section structure as metadata. After filtering out overly short paragraphs and cleaning special characters, we complete the context by incorporating the paragraph text and appending the relevant section titles. A complete example is provided in Figure \ref{fig:intro}. 

There are existing datasets that contain human-generated questions also use Wikipedia as context. 
As introduced in Section \ref{sec:related}, HotpotQA~\citep{yang2018hotpotqa} is a human-labeled multi-hop QA dataset, where workers on Mechanical Turk generate questions based on multiple evidence paragraphs from Wikipedia. In contrast, TriviaQA~\citep{JoshiTriviaQA2017} is a QA dataset compiled by trivia enthusiasts, where the questions are created first, and relevant evidence from Wikipedia is subsequently identified to support the answers. Although both datasets incorporate Wikipedia knowledge and human-generated questions, they represent different workflows for introducing evidence: HotpotQA presents context first, with humans generating questions based on it, while TriviaQA starts with the questions and then seeks evidence.

In our work, we adopt an automatic, context-dependent approach to generate questions, similar to HotpotQA's context-first methodology.
We explore the characteristics of LLM-generated questions by comparing them to these human-generated questions, highlighting differences in question characteristics and quality criteria.

\subsection{Answer-irrelevant metrics}
\label{sec:irrmet}

The use of LLMs presents new opportunities for more comprehensive evaluation. While previous work~\citep{es2023ragas} has addressed aspects such as faithfulness, answer relevance, and context relevance, we extend this by introducing finer-grained statistical measurements and new evaluation methods. Specifically, we focus on two groups of criteria, totaling six. The first three assess the questions themselves, while the second involves evaluating generated answers as a proxy for assessing the quality of the underlying questions.

\paragraph{Question type.}
For humans, the choice of which question to ask is subjective. Here, we explore the types of questions LLMs can generate without additional constraints and compare them with human preferences. Previous work~\citep{yang2018hotpotqa} introduced heuristic rules, categorizing questions based on their starting words. In this work, we extend this by incorporating a LLM-based classification to analyze question types across ten manually defined categories. These categories were developed by applying inductive coding on a mix of questions from HotpotQA, TriviaQA, and our generation results. Specifically, we apply automatic balanced grouping using LLMs for insights and further refined by human review. Below we provide detailed definitions for these question types.

\begin{enumerate}
    \item \textbf{Identity and Attribution Questions}: These inquiries focus on identifying a person or entity responsible for an action or associated with a work. They tend to ask "Who...?" or refer to persons or origins related to a context.
    \item \textbf{Which/What-Based General Knowledge Questions}: This group contains questions that start with "Which" or "What" and inquire about general knowledge, often requiring a selection from a set or identification of a type/category.
    \item \textbf{Location-Based Questions}: These questions focus on identifying a geographic location or specific place where something is based or occurs.
    \item \textbf{Classification and Categorization Questions}: These inquiries request the classification or categorical identity of entities or things, often seeking to place an item within a broader group or category. 
    \item \textbf{Specific Fact and Figure Questions}: These questions request a specific quantitative or qualitative fact. They are straightforward and seek concrete data or a precise answer, often involving numbers or specific details.
    \item \textbf{Comparison and Selection Questions}: Questions in this group involve comparing two entities to determine which one holds a particular status or characteristic, often using formats like "Between X and Y, who/which is...?"
    \item \textbf{Verification/Affirmation Questions}: These questions ask for confirmation about the equivalence or relationship between two or more entities. They often use formats like "Are...?" or "Which...?"
    \item \textbf{Descriptive/Characterization Questions}: These questions seek an explanation or characterization of entities, often requiring a description of how or why something is the way it is, involving traits or actions.
    \item \textbf{Event/Outcome Questions}: These questions inquire about the outcome of specific events or actions, focusing on consequences or results. They often address changes, damages, or effects.
    \item \textbf{Sequential/Ordering/Causation Questions}: These questions require identifying a sequence, comparison, or causation among entities, often using terms like "first," "before," "between," etc.
\end{enumerate}

While we prompt the LLM to use the category "Others" when a question does not fit any of the above categories, we find that over 99.9\% of both human- and LLM-generated questions can be classified into the existing categories. Besides, although we aim for orthogonal category definitions to simplify classification, some overlap remains. To address this, we manually refine category descriptions to ensure that, in most cases, one category provides the best fit.

\paragraph{Question length.}
Length is the most straightforward statistical measure of generated questions. In this work, we mainly measure word count numbers. In addition to directly comparing question lengths between human-generated and LLM-generated datasets, we also examine how length relates to question content, such as question type. Analyzing average lengths across different question types helps clarify systematic differences between human and LLM generation.

\paragraph{Context coverage.}
A key attribute of QG is whether the questions are relevant to the context. Additionally, since the context often includes multiple sentences, a question may either address a single fact from one sentence or require reasoning across multiple sentences. While previous work~\citep{es2023ragas} introduced a sentence-level measurement using prompts, we extend this work in two ways. 
First, we also investigate word-level context coverage for finer granularity. 
Second, we analyze which specific parts of the context LLMs tend to focus on during generation.
We first split the context into sentences and number them in a list. Then we use the following prompt to ask LLM to discover which particular sentences are relevant to the generated question.
We find that QG does not follow similar positional bias in QA discussed in previous work~\cite{saito2024answerinvestigatingpositionalbias}.

\begin{prompt}
\textbf{Context coverage.} Select the minimal set of context sentences most relevant to answering the question. 
You need to choose at least one sentence and can select multiple sentences.
Output only the sentence numbers of these sentences in a comma-separated list on a single line without any additional text.
\end{prompt}


\subsection{Answer-relevant metrics}
\label{sec:ar}
The quality of the questions are often reflected in their answers. In addition to asking LLM to generate the question, we also use LLM to generate answers according to the context and rate the quality of the answer. Specifically, we adopt the star-rating approach from the previous work~\citep{wang2023chatgpt} and use the following prompts.

\begin{prompt}
\textbf{Answer generation.} You are to generate a short answer based on the following question and an optional supporting fact.
\end{prompt}
\vspace{-0.3cm}
\begin{prompt}
\textbf{Answer rating.} You are to rate the following answer to a question, taking into account any optional supporting facts provided. 
    
Assign a rating from 0 to 5 based on the criteria below:

0: No answer or completely irrelevant

1: Significantly incorrect or incomplete

2: Partially correct; major inaccuracies or omissions

3: Correct but lacks depth; minimal detail

4: Mostly correct; minor errors; includes relevant details

5: Fully accurate and detailed; clear and comprehensive

Your response should consist of two lines:
The rating from 0 to 5.
A brief justification for your rating.
\end{prompt}

\paragraph{Answerability.}
A key quality criterion for a question is whether it can be precisely answered given specific knowledge. In concrete terms, a generated question should be answerable when the context is provided. To assess the answerability of a question, we prompt the LLM to generate an answer using the given context as input. Since answer correctness is also evaluated based on the same context, in most cases, the generated questions are answerable.

\paragraph{Uncommonness.}
LLMs are trained on widely available common knowledge from the Internet. As a result, even when the context is not explicitly provided, the LLM may still answer the question. Evaluating the answer quality without giving the LLM the context thus becomes a way to assess the question's uncommonness relative to its pretraining data. Compared to the answerability evaluation, the key difference is that the context is omitted during answer generation, while other factors remain unchanged. As we demonstrate later, removing the context significantly reduces the answer quality.
This also suggests that the generated questions are valuable for evaluating RAG systems or for automated hallucination testing.

\paragraph{Required answer length.}
In addition to question length, the required answer length serves as a more effective measure of how much information is asked. However, due to the nature of generative models, generated answers tend to be longer and include more details. To filter out unnecessary information from the answers generated with context, we implement two strategies to measure the essential answer length. 
First, we explicitly prompt the model to provide a shortest text answer (``Provide a very concise answer without repeating the question.''). 
Second, we set a word limit and instruct the model to generate answers within that limit (``Please ensure that your answer contains no more than X words.''). 
In practice, we set the word limit to 1, 2, 3, 4, and 8.
We then evaluate these new answers to determine if they can achieve the same quality rating with fewer words. If a shorter version of the answer receives the same rating as the original, we treat the shorter one as sufficient. Our approach significantly reduces answer length, with the second strategy generally proving more effective. This provides a direct measure of the information needed to answer the question.


\section{Experiments}
We present the evaluation results using four representative LLMs, two open-source and two closed-source. For closed-source models, we picked the GPT-4o-2024-08-06 by ChatGPT and Claude-3.7-Sonnet by Anthropic; 
and for the open-source models representatives we used Llama-3.3-70B-Instruct~\citep{dubey2024llama} and DeepSeek-V3~\citep{deepseekai2024deepseekv3technicalreport} with temperature set to 0 and all other parameters set to default.
We run open-source models using TogetherAI and the complete evaluation uses around 70k chat completion calls, which takes around 2 hours to run.

For human questions, we compare TriviaQA under Apache 2.0, and HotpotQA under CC BY-SA 4.0.
For LLM-based QA, we use the context text sampled from the WikiText dataset~\citep{merity2016pointer} under CC-BY-AS-3.0 license.
Each LLM generates 1,024 questions with a shared set of 256 sampled Wiki contexts (N=4). 
To control the influence of prompt phrasing, we add two different paraphrased prompts for QG to repeat the evaluation and the observations are consistent as shown in Appendix \ref{sec:diffprompt}.
For metrics that require the LLM as a judge, we use GPT-4o for consistency. 
With six metrics presented in Section \ref{sec:methodology}, we evaluate LLM-based QG and answer five key questions revealing the generation preference. 





\paragraph{What types of question do LLMs ask?} 

Classified by the predefined question types in Section \ref{sec:irrmet}, Table \ref{tab:qtype} compares LLM-generated questions with those from human datasets.

\begin{table}[h]
\small
\resizebox{\textwidth}{!}{%
\centering
\begin{tabular}{lrrrrrr}
                                   \hline
                                  & TriviaQA & HotpotQA & Llama-3.3 & DeepSeek-V3 & Claude-3.7 & GPT-4o \\ \hline
T1 Identity/Attribution           & 34.2     & 39.7     & 12.9  & 9.3      & 14.8   & 11.2 \\
T2 General Knowledge              & 34.5     & 15.0     & 7.7   & 6.1      & 8.8    & 7.1  \\
T3 Location                       & 12.2     & 14.3     & 5.2   & 4.1      & 3.8    & 3.7  \\
T4 Classification/Categorization  & 4.3      & 2.7      & 3.0   & 1.4      & 2.5    & 2.1  \\
T5 Specific Fact/Figure           & 10.5     & 9.5      & 30.5  & 18.3     & 28.6   & 21.0 \\ \cline{1-1}
T6 Comparison/Selection           & 0.1      & 6.7      & 1.0   & 0.8      & 0.9    & 1.0  \\
T7 Verification/Affirmation       & 0.1      & 6.5      & 0.1   & 0.0      & 0.0    & 0.1  \\ \cline{1-1}
T8 Descriptive/Characterization   & 3.0      & 1.5      & 27.1  & 44.7     & 29.4   & 37.1 \\
T9 Event/Outcome                  & 0.2      & 0.8      & 11.9  & 13.8     & 9.6    & 14.1 \\
T10 Sequential/Ordering/Causation & 0.9      & 3.2      & 0.6   & 1.7      & 1.6    & 2.0  \\ \cline{1-1}
Others (failed to classify)       & 0.0      & 0.0      & 0.0   & 0.0      & 0.0    & 0.0  \\ \hline
\end{tabular}%
}
\caption{\label{tab:qtype} 
Percentage of different question types across different datasets. Three groups are formed according to the comparison results. 
}
\end{table}

Based on the comparison, we categorize the question types into three groups and discuss the findings.
\begin{itemize}
    \item The first group (T1–T5) centers on factual checking, with substantial representation of both human- and LLM-generated questions. \textbf{LLMs exhibit a strong tendency to ask about specific facts and figures}, likely reflecting the prevalence of numerical data and named entities in their training corpus. In contrast, human-generated questions more frequently target identities and attributions—often concerning other people—highlighting the inherently social nature of human inquiry.
    \item Questions in the second group (T6-T7) require reasoning across multiple facts from the context. HotpotQA, designed for multi-hop questions, contains a higher number of questions in this category.
    \item The third group (T8-T10) involves questions that require more descriptive answers. \textbf{We observe a strong preference for asking descriptive questions in LLMs.} As we discuss later, this preference also leads to longer answers.
\end{itemize}
When comparing to human generated questions, LLM-generated questions exhibit consistent patterns for question type preferences. 
Among the four models, while the general distributions are quite similar, DeepSeek and ChatGPT have an even stronger preference for descriptive questions.


\paragraph{How long are the generated questions?}
From Table \ref{tab:qlen}, we observe that 
while overall question lengths are quite similar to be around 18 words, different LLMs tend to exhibit distinct preferences for length, with Llama generating longer questions. 
Additionally, the length of LLM-generated questions consistently shows lower variation.
We also measure the question length for different type of questions generated by GPT-4o. The result suggests that despite the difference in question type, the question length is quite consistent.

\begin{table}[h]
\centering
\begin{tabular}{ll}
\\ \hline
                       & Q Len ± std   \\
\hline
TriviaQA               & 14.0 ± 6.5    \\
HotpotQA               & 18.0 ± 10.0   \\
Llama-3.3              & 23.6 ± 5.6    \\
DeepSeek-V3            & 17.6 ± 4.0    \\ 
Claude-3.7             & 15.3 ± 3.6    \\ 
GPT-4o                 & 16.7 ± 4.0    \\ \cline{1-1}
GPT-4o T1-T5 (45.8\%)  & 16.6 ± 4.2    \\
GPT-4o T6-T7 (0.8\%)   & 20.6 ± 4.3    \\
GPT-4o T8-T10 (53.4\%) & 16.7 ± 3.7    \\
\hline
\end{tabular}
\caption{\label{tab:qlen} Question length statistics (word count) 
}
\end{table}

\paragraph{How much context is used and which part?}
Table \ref{tab:qcov} presents the context coverage analysis. 
TriviaQA uses a different format of source paragraph for its labeling, thus is excluded from this comparison.
We observe that human-generated questions tend to cover more of the context, with consistent results across both sentence-level and word-level measurements. 
Among different LLMs, DeepSeek achieves the highest coverage that is close to human generated questions.

\begin{table}[h] 
\centering
\begin{tabular}{lrrrr}
                  &                               &                  &                            &                  \\ \hline
                  & \# Words in Ctx               & \% Covered Words & \# Sents in Ctx            & \% Covered Sents \\ \hline
HotpotQA          & 139.1 ± 53.3                  & 33.6 ± 17.5      & 6.5 ± 2.5                  & 31.1 ± 15.6      \\ \hline
Llama-3.3         & \multirow{4}{*}{140.3 ± 56.1} & 29.8 ± 17.6      & \multirow{4}{*}{5.6 ± 2.7} & 28.5 ± 15.1      \\
DeepSeek-V3       &                               & 32.0 ± 18.2      &                            & 30.4 ± 15.9      \\
Claude-3.7 &                               & 28.8 ± 16.9      &                            & 26.6 ± 13.8      \\
GPT-4o            &                               & 28.5 ± 17.2      &                            & 26.7 ± 13.9      \\ \hline 

\end{tabular}
\caption{\label{tab:qcov} Context coverage in word and sentence level. All LLM-generated questions use the same subsample of WikiText as context, thus the number of words and sentences in context is the same.}
\end{table}

\begin{figure}[h]
    \centering
    \includegraphics[width=\textwidth]{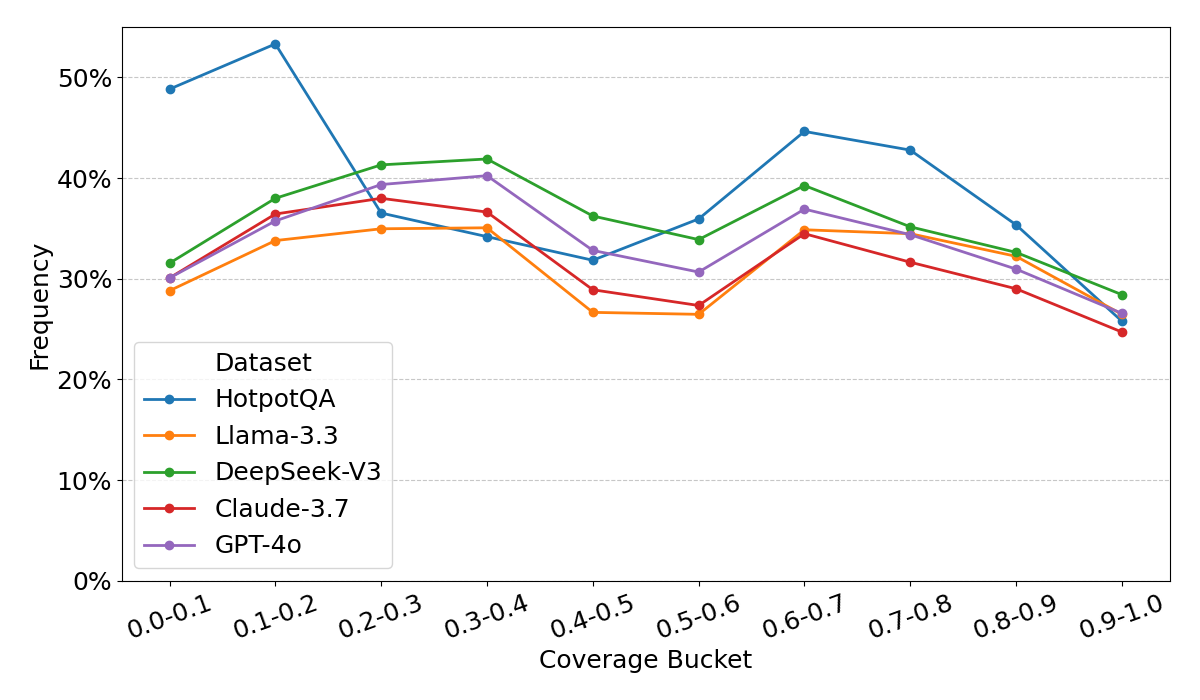}
    \caption{\label{fig:qcov} Context coverage region analysis. Ten region buckets are created and access frequency is provided. For example, 30\% access within 0.0-0.1 bucket suggests 30\% of the questions refer to the text in the beginning 10\% of the context.
    }
\end{figure}

Figure \ref{fig:qcov} further investigates which regions of the context LLMs prefer to focus on when generating questions. We observe that human-generated questions tend to focus on the beginning of the context, likely due to the labeling process. 
LLM-generated questions exhibit a more balanced distribution, with a slight decrease in focus at both ends. This suggests that \textbf{LLM-based QG displays a different positional focus compared to QA}~\cite{saito2024answerinvestigatingpositionalbias}.
Such positional focus is highly consistent across different LLMs, which possibly reflects about how information is distributed in the context.

\begin{figure}[h]
    \centering
    \includegraphics[width=\textwidth]{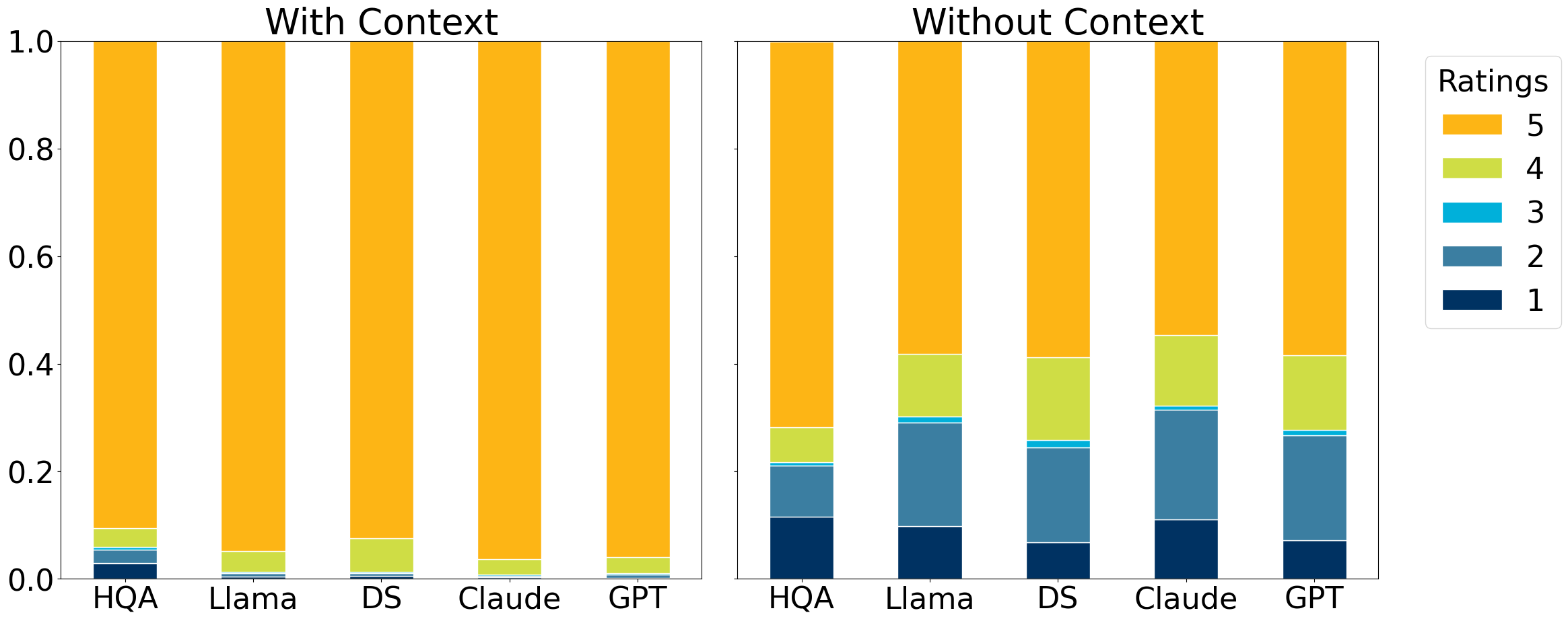}
    \caption{Answer ratings distribution with and w/o context. No zero-rating answers are found. HQA denotes the HotpotQA dataset, Llama denotes Llama-3.3-70B, DS denotes DeepSeek-V3-0324, Claude denotes Claude-3.7-Sonnet, GPT denotes GPT-4o. TriviaQA is not included due to its different context setup.}
    \label{fig:ans}
\end{figure}

\paragraph{Are generated questions answerable with and w/o context?}
As mentioned in Section \ref{sec:ar}, we use the LLM as a judge to rate answers for generated questions.
To verify the human alignment in the rating task, from a sample of 300 manual annotations, the GPT-4o judge achieves a Pearson correlation of 0.76.
By combining answer generation and rating, Figure \ref{fig:ans} shows the rating distribution. 

We observe that with context, the LLM typically generates satisfactory answers, meeting expectations. Also, performance declines when context is not provided. Notably, about one-fourth of the generated questions cannot be properly answered. \textbf{Questions generated by LLMs show an even higher rate of uncommon questions compared to the human-constructed HotpotQA dataset}, highlighting their potential for automatic testing in RAG systems or detecting model hallucinations.
While LLMs in general exhibits similar ability in generating uncommon questions, Claude yields slightly more uncommon questions than others.
Besides, for GPT-4o generated questions, although we use the same GPT-4o model for judging answer quality, there isn't a clear difference when compared with judging questions generated by other LLMs.

\paragraph{How much information is needed to answer the question?}
As shown in Table \ref{tab:shorter}, LLM-generated answers are generally much longer than human-labeled golden answers, likely due to the nature of generative models. 
We also find that DeepSeek and GPT-4o tend to generate even longer answers compared with others.
To better measure the amount of information required, we compress the LLM-generated answers to obtain a minimum-length version that maintains the same rating using methods mentioned in Section \ref{sec:ar}.
Appendix \ref{sec:effective_answer_shortening} provides a detailed analysis on the answer shortening procedure.
While this compression does reduce answer length, \textbf{LLM-generated questions still require significantly longer answers}. Further analysis, breaking down answer length by question type, reveals that comparitive and descriptive questions are the primary reason for longer answers. We can observe that T6-T10 questions generated by GPT-4o indeed exhibit longer shortened answers and larger variance.

\begin{table}[h]
\centering
\begin{tabular}{lrrr}
\\ \hline
                & Original Answer Length & Shortened Answer Length &  \\ \hline
TriviaQA-golden & 2.0 ± 1.8       & -                       &  \\
HotpotQA-golden & 2.2 ± 1.8       & -                       &  \\
Llama-3.3           & 24.8 ± 15.0     & 7.3 ± 12.6   
&  \\
DeepSeek-V3           & 35.0 ± 18.7     & 13.7 ± 19.1               &  \\
Claude-3.7             & 24.4 ± 14.2     & 7.5 ± 11.8                &  \\
GPT-4o          & 30.7 ± 17.2     & 10.4 ± 15.7                &  \\
GPT-4o T1-T5 (45.8\%) & 21.6 ± 10.3      & 6.8 ± 10.0                &  \\
GPT-4o T6-T7 (0.8\%) & 39.8 ± 19.4      & 16.5 ± 16.9                 &  \\
GPT-4o T8-T10 (53.4\%) & 38.4 ± 18.0     & 13.9 ± 19.6              &  \\ \hline
\end{tabular}
\caption{\label{tab:shorter} Answer length statistics and shortening experiment. LLM answers are generated with context. We observe that T6-T10 have both longer original answers and longer shortened answers. }
\end{table}

\section{Discussion}



With the growing use of LLMs in human-centric applications, it is increasingly important to understand how LLM-generated questions differ from human-authored ones.
In this work, we identify distinct characteristics of LLM-generated questions across various models. Notably, LLMs consistently favor descriptive questions requiring longer answers, exhibit less positional bias toward the beginning of the context compared to humans, and are capable of generating unanswerable questions, which can be used for automatic evaluation in tasks like RAG and hallucination detection.
Despite these insights, several limitations remain, including the need to incorporate reasoning models, assess the impact of prompt design with preferences, include more human annotations, and extend evaluation to additional downstream tasks.

This paper investigates LLM-generated questions using representative models in default settings. However, many potential variations remain unexplored. Future research could expand to include additional models and employ more sophisticated sampling algorithms.
For example, while we exclude reasoning models such as ChatGPT-o3 or DeepSeek-R1 from our experiment candidates, it might be interesting to explore how reasoning procedure affects the QG preference, as well as finding relevant indicators from the thinking procedure.

While we evaluated prompts without generation preferences in three variations, which is sufficient for our observations, in practice, users may adjust prompts for more targeted question generation. The goal of this work is to highlight that, similar to LLM behavior analysis in QA, there are significant differences between human-generated and LLM-generated questions in QG tasks. The findings in this paper serve as a case study and provide insights for future prompt engineering.

Due to resource constraints, we only evaluated human alignment on a sampled dataset and did not incorporate large-scale human annotation using public resources such as Mechanical Turk. Future studies that integrate more extensive human annotations and cross-checking could yield more accurate quantitative measurements of human-LLM differences.

Although this paper focuses on general QG settings, it would also be valuable to explore LLM-based QG for specific downstream tasks, such as the RAG system evaluation~\citep{es2023ragas} or hallucination detection. We expect future research to provide more detailed application-specific analysis. Similarly, exploring QG using context from specialized domains, such as finance or medical texts, could enhance the application of QG in domain-specific education.

\bibliography{custom}
\bibliographystyle{plainnat}
\clearpage
\appendix
\section*{APPENDIX}

\section{Effect of prompt wording}
\label{sec:diffprompt}

Here we observe how prompt wording affect question generation. To rule out wording effects in our results, we reran question generation with the following paraphrased (by GPT-4o) prompts with similar meaning:

\begin{prompt}
\textbf{Question generation V1.} Generate [N] self-contained questions based on the following content in an ordered list.
\end{prompt}

\vspace{-0.2cm}

\begin{prompt}
\textbf{Question generation V2.} Create [N] questions based on the following content in an ordered list.
\end{prompt}

\vspace{-0.2cm}

\begin{prompt}
\textbf{Question generation V3.} Reference exclusively the content below to craft [N] independent questions. Format your output as an ordered list.
\end{prompt}


\subsection{Question types}

We can observe here that with different prompts, the distribution of different question types remains stable, consistently showing that the LLMs prefer descriptive questions.

\begin{table*}[h]
\small
\resizebox{\textwidth}{!}{%
\centering
\begin{tabular}{lrrrrrrrrrrrr}
\hline
                                  & \multicolumn{3}{c}{Llama-3.3} & \multicolumn{3}{c}{DeepSeek-V3} & \multicolumn{3}{c}{Claude-3.7} & \multicolumn{3}{c}{GPT-4o} \\ \hline
                                  & v1       & v2       & v3      & v1        & v2       & v3       & v1       & v2       & v3       & v1      & v2      & v3     \\ \hline
T1 Identity/Attribution           & 12.9     & 13.3     & 12.5    & 9.3       & 8.0      & 7.9      & 14.8     & 15.6     & 13.5     & 11.2    & 12.2    & 12.7   \\
T2 General Knowledge              & 7.7      & 8.8      & 8.4     & 6.1       & 7.5      & 5.1      & 8.8      & 9.4      & 9.3      & 7.1     & 7.8     & 8.5    \\
T3 Location                       & 5.2      & 5.1      & 5.7     & 4.1       & 4.4      & 2.8      & 3.8      & 5.1      & 4.4      & 3.7     & 4.4     & 4.7    \\
T4 Classification/Categorization  & 3.0      & 2.8      & 3.1     & 1.4       & 1.8      & 1.4      & 2.5      & 2.6      & 1.7      & 2.1     & 1.8     & 2.0    \\
T5 Specific Fact/Figure           & 30.5     & 31.9     & 33.8    & 18.3      & 17.7     & 18.1     & 28.6     & 30.2     & 28.2     & 21.0    & 25.2    & 25.6   \\ \cline{1-1}
T6 Comparison/Selection           & 1.0      & 0.5      & 0.6     & 0.8       & 1.0      & 0.9      & 0.9      & 0.5      & 0.8      & 1.0     & 0.8     & 0.8    \\
T7 Verification/Affirmation       & 0.1      & 0.1      & 0.0     & 0.0       & 0.0      & 0.1      & 0.0      & 0.1      & 0.0      & 0.1     & 0.1     & 0.1    \\ \cline{1-1}
T8 Descriptive/Characterization   & 27.1     & 25.9     & 23.9    & 44.7      & 45.6     & 49.3     & 29.4     & 26.1     & 31.0     & 37.1    & 33.8    & 32.0   \\
T9 Event/Outcome                  & 11.9     & 11.0     & 11.2    & 13.8      & 12.4     & 13.0     & 9.6      & 9.6      & 10.0     & 14.7    & 12.8    & 12.6   \\
T10 Sequential/Ordering/Causation & 0.6      & 0.6      & 0.8     & 1.7       & 1.7      & 1.4      & 1.6      & 0.9      & 1.3      & 2.0     & 1.2     & 1.1    \\ \cline{1-1}
Others (failed to classify)       & 0.0      & 0.0      & 0.0     & 0.0       & 0.0      & 0.1      & 0.0      & 0.0      & 0.0      & 0.0     & 0.0     & 0.0    \\ \hline

\end{tabular}%
}

\caption{\label{tab:qtype_prompt_ver} Percentage of different question types across different prompt versions. Three groups are formed according to the comparison results.}
\end{table*}

\subsection{Question length}

We can observe here that with different prompts, the variance of LLM-generated question length is consistently lower than human questions.

\begin{table}[h]
\centering
\begin{tabular}{lrrrrr}
\scriptsize
\\ \hline
                       && Q Len ± std  \\
\hline
                       & v1 & v2 & v3  \\
\hline
TriviaQA               & 14.0 ± 6.5 & - & -  \\
HotpotQA               & 18.0 ± 10.0 & - & -  \\
Llama-3.3-70B                  & 18.5 ± 4.6 & 17.4 ± 4.4 & 17.3 ± 4.4   \\
DeepSeek-V3-0324                  & 17.6 ± 4.0 & 16.2 ± 3.7 & 18.0 ± 4.1   \\ 
Claude-3.7-Sonnet                 & 15.3 ± 3.6 & 14.2 ± 3.8 & 15.2 ± 3.7   \\ 
GPT-4o                 & 16.7 ± 4.0 & 15.1 ± 3.9 & 15.4 ± 3.9   \\ \cline{1-1}
GPT-4o T1-T5 (58.0\%)  & 16.3 ± 4.0 & 14.6 ± 3.9 & 14.8 ± 3.8   \\
GPT-4o T6-T7 (0.8\%)   & 16.8 ± 3.8 & 15.4 ± 3.5 & 15.5 ± 3.7   \\
GPT-4o T8-T10 (41.2\%) & 16.7 ± 4.0 & 15.1 ± 4.0 & 15.5 ± 3.9   \\
\hline
\end{tabular}
\caption{\label{tab:qlen_version} Question length statistics (word count)}
\end{table}

\section{Effectiveness of answer shortening}
\label{sec:effective_answer_shortening}

\begin{figure}[h]
    \centering
    \begin{subfigure}[b]{0.49\textwidth}
        \centering
        \includegraphics[width=\textwidth]{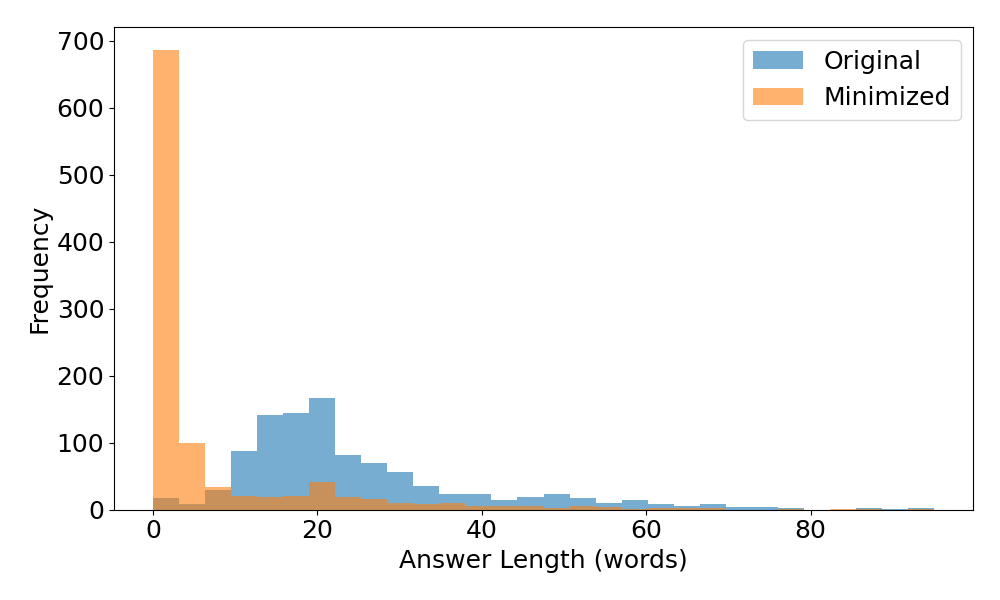}
        \caption{Llama-3.3}
        \label{fig:ans1}
    \end{subfigure}
    \hfill
    \begin{subfigure}[b]{0.49\textwidth}
        \centering
        \includegraphics[width=\textwidth]{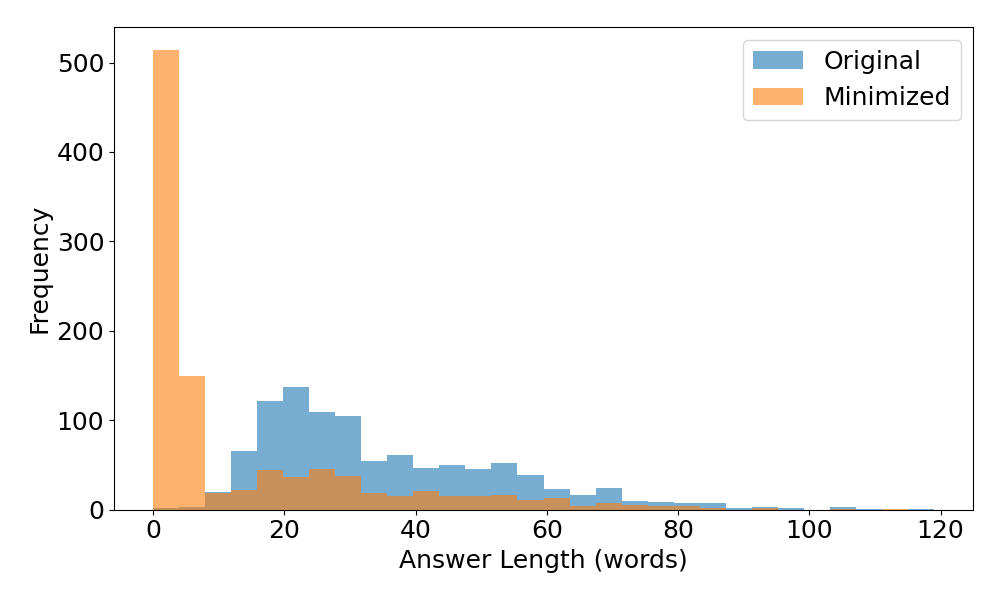}
        \caption{DeepSeek-V3}
        \label{fig:ans2}
    \end{subfigure}
    \vskip\baselineskip
    \begin{subfigure}[b]{0.49\textwidth}
        \centering
        \includegraphics[width=\textwidth]{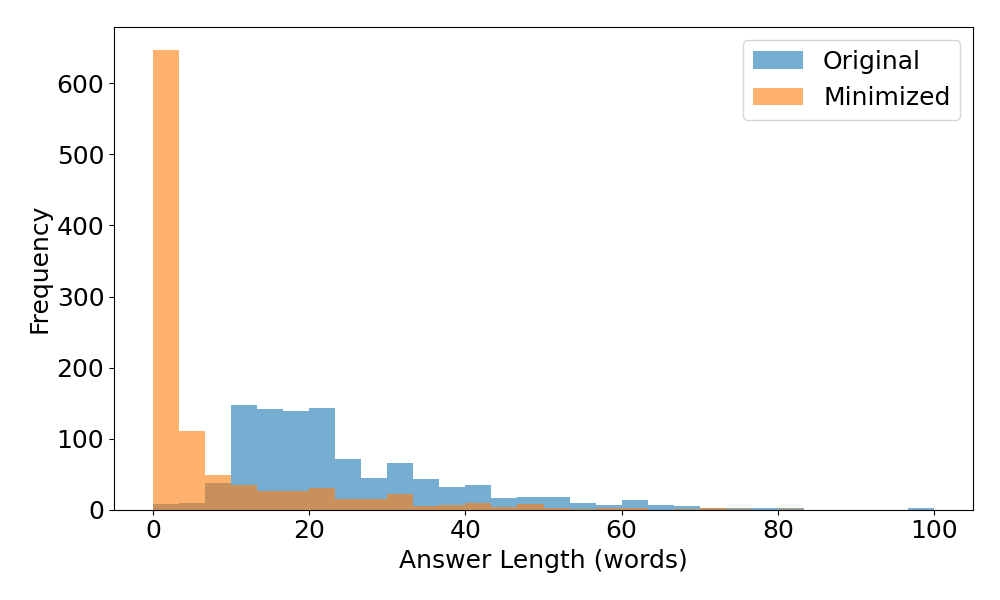}
        \caption{Claude-3.7}
        \label{fig:ans3}
    \end{subfigure}
    \hfill
    \begin{subfigure}[b]{0.49\textwidth}
        \centering
        \includegraphics[width=\textwidth]{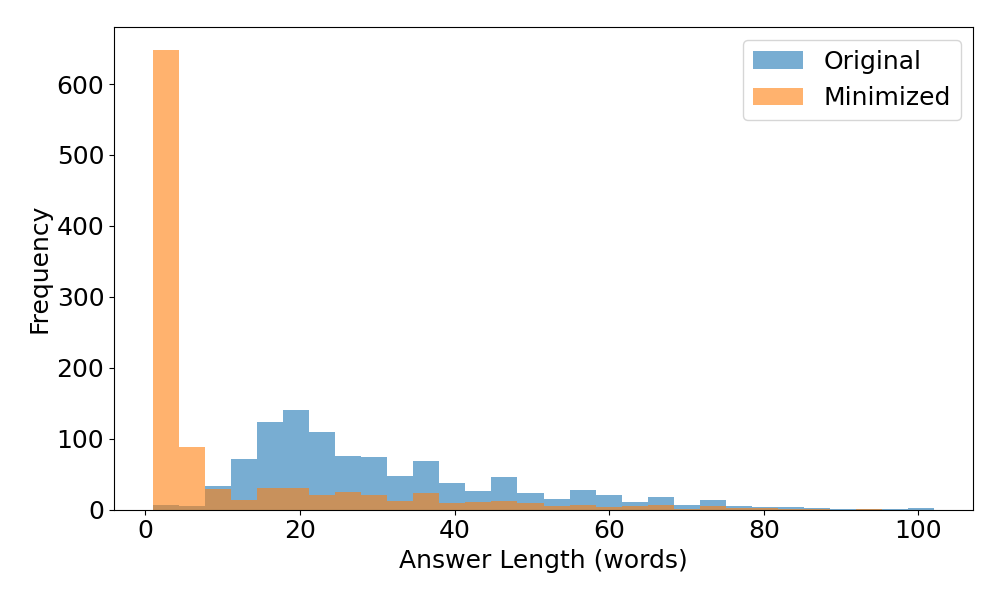}
        \caption{GPT-4o}
        \label{fig:ans4}
    \end{subfigure}
    \caption{Shortened answer length distribution vs. original answer length distribution across questions generated by four LLMs.}
    \label{fig:ans_all}
\end{figure}

The effectiveness of answer length shortening can be observed in Figure \ref{fig:ans_all}, where we compare the distribution shifts of answer lengths when we shorten the answers across datasets generated by four models. As we can see the shortened answer length distribution generally shifted to the left, as answers got distilled down into as few as 1 word. More interestingly, in all four instances there is a notable long tail remaining in the distribution, where a significant fraction of LLM-generated questions cannot be fully and correctly answered in only a few words.

\begin{figure}[h]
    \centering
    \begin{subfigure}[b]{0.32\textwidth}
        \centering
        \includegraphics[width=\textwidth]{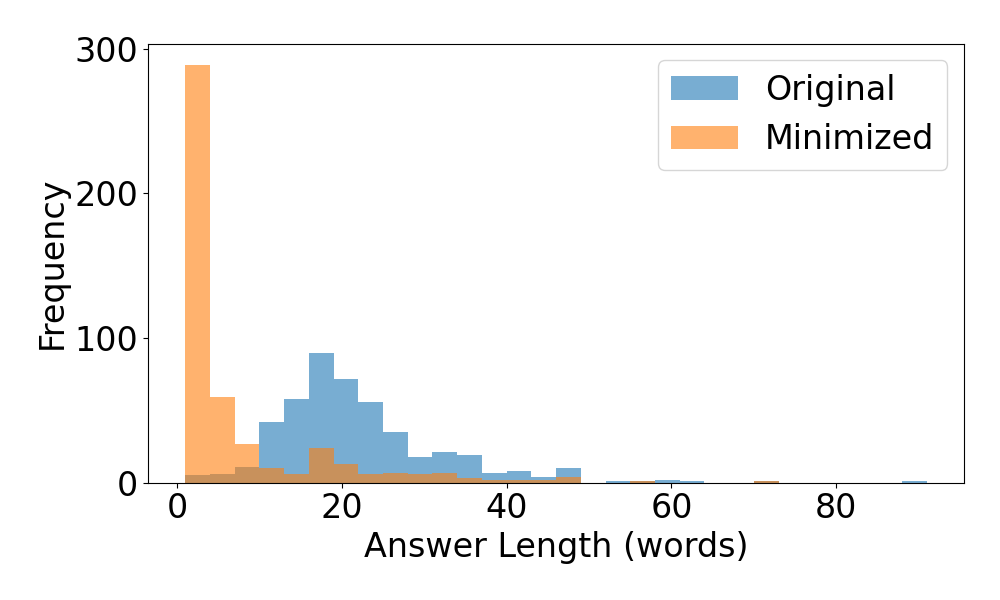}
        \caption{GPT-4o T1-T5}
        \label{fig:panel1}
    \end{subfigure}
    \hfill
    \begin{subfigure}[b]{0.32\textwidth}
        \centering
        \includegraphics[width=\textwidth]{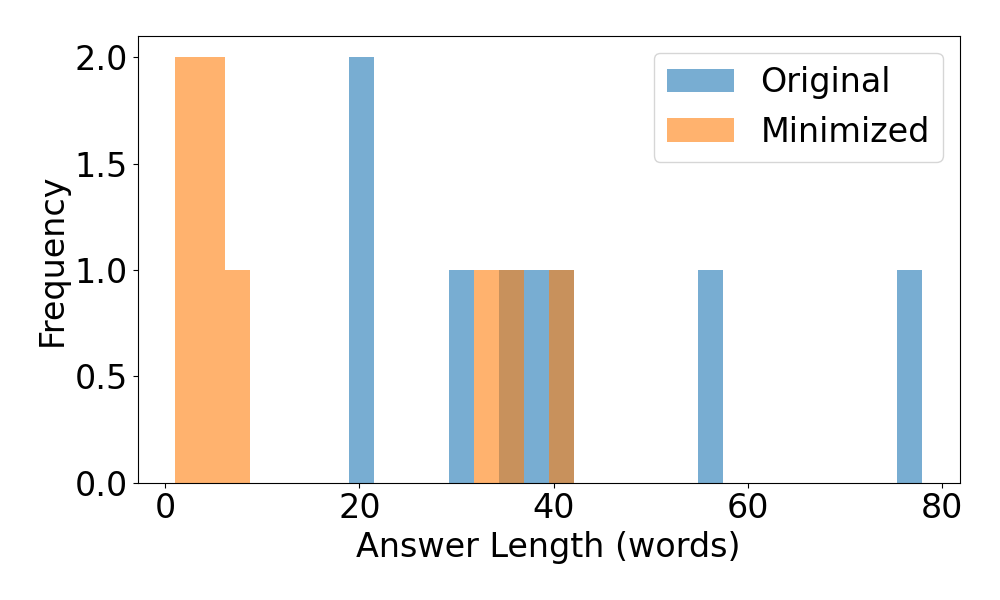}
        \caption{GPT-4o T6-T7}
        \label{fig:panel2}
    \end{subfigure}
    \hfill
    \begin{subfigure}[b]{0.32\textwidth}
        \centering
        \includegraphics[width=\textwidth]{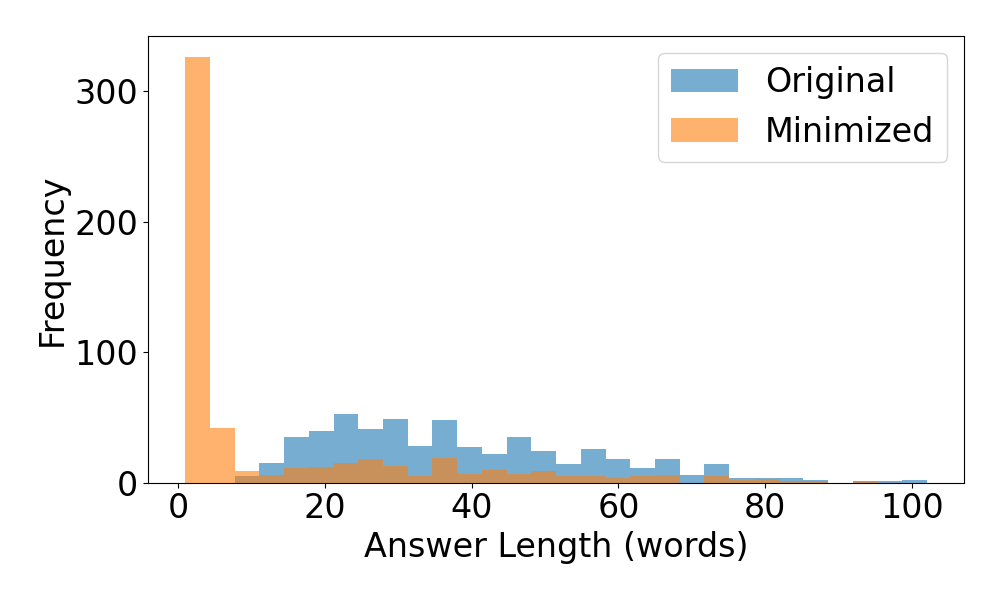}
        \caption{GPT-4o T8-T10}
        \label{fig:panel3}
    \end{subfigure}
    \caption{Shortened Answer Length Distribution vs. Original Answer Length Distribution across question types.}
    \label{fig:three_panel_row}
\end{figure}

In Figure \ref{fig:panel3} we compare the distribution shift before and after the answer shortening for the dataset generated by GPT-4o, across different question types. We observe that the original answers for T6-T10 is generally longer and more varied when compared to T1-T5. After shortening, the required answer length in all groups sees a shift to the left, with T6-T10 preserving the most of its long answers.



\section{Code Availability}
\label{appendix:code}

The source code supporting the findings of this study is publicly available at our GitHub repository:

\begin{center}
    \href{https://github.com/CoLearn-Dev/llmqg}{\texttt{https://github.com/CoLearn-Dev/llmqg}}
\end{center}

\end{document}